\pdfoutput=1

\documentclass[11pt]{article}

\usepackage{ACL2023}
\usepackage{bbm}
\usepackage{graphicx}
\usepackage{amsmath}
\usepackage{amssymb}
\usepackage{pifont}
\usepackage{times}
\usepackage{latexsym}
\usepackage{enumitem}
\usepackage{xspace}

\usepackage{listings}
\usepackage[T1]{fontenc}

\usepackage[utf8]{inputenc}

\usepackage{microtype}

\usepackage{inconsolata}

\usepackage{booktabs}
\usepackage{multirow}

\title{\textsc{taxi}: Evaluating Categorical Knowledge Editing for Language Models}

\newcommand{\asu}{$^\vartriangle$}
\newcommand{\uva}{$^\heartsuit$}
\newcommand{\MIT}{$^\diamondsuit$}

\newcommand{\bspace}{\hspace{1em}}

\author{
    Derek Powell\asu\bspace
    Walter Gerych\MIT\bspace
    Thomas Hartvigsen\uva\\
    \asu Arizona State University\bspace \MIT MIT\bspace \uva University of Virginia\\
    {\small \texttt{dmpowell@asu.edu,wgerych@mit.edu,hartvigsen@virginia.edu}}
}

\begin{document}
\maketitle

\begin{abstract}
Humans rarely learn one fact in isolation. Instead, learning a new fact induces knowledge of other facts about the world. For example, in learning a \texttt{korat} is a type of \texttt{cat}, you also infer it \texttt{is a mammal} and \texttt{has claws}, ensuring your model of the world is \textit{consistent}. Knowledge editing aims to inject new facts into language models to improve their factuality, but current benchmarks fail to evaluate consistency, which is critical to ensure efficient, accurate, and generalizable edits. We manually create \textsc{taxi}, a new benchmark dataset specifically created to evaluate consistency in categorical knowledge edits. \textsc{taxi} contains 11,120 multiple-choice queries for 976 edits spanning 41 categories (e.g., Dogs), 164 subjects (e.g., Labrador), and 183 properties (e.g., is a mammal). We then use \textsc{taxi} to evaluate popular editors' categorical consistency, measuring how often editing a subject's category appropriately edits its properties. We find that 1) the editors achieve marginal, yet non-random consistency, 2) their consistency far underperforms human baselines, and 3) consistency is more achievable when editing atypical subjects.\footnote{\url{https://github.com/derekpowell/taxi}}
\end{abstract}

\section{Introduction}

Many recent works aim to edit the memorized factual associations encoded in Large Language Models (LLMs) \cite{cohen2023evaluating, dai.etal2022, hartvigsen2023aging, huang2023transformer, mazzia2023survey, meng.etal2022,  meng2022locating, mitchell2021fast, mitchell2021fast, mitchell.etal2022a, tan2024massive, wang2023knowledge, zhong2023mquake}.
If effective, such techniques could offer a transparent and explainable means of updating out-of-date information; correcting biased, offensive, or inaccurate outputs; deleting or obscuring unwanted information to support privacy; and helping to personalize models.

\begin{figure}[t]
    \centering
    \includegraphics[width=0.9\linewidth]{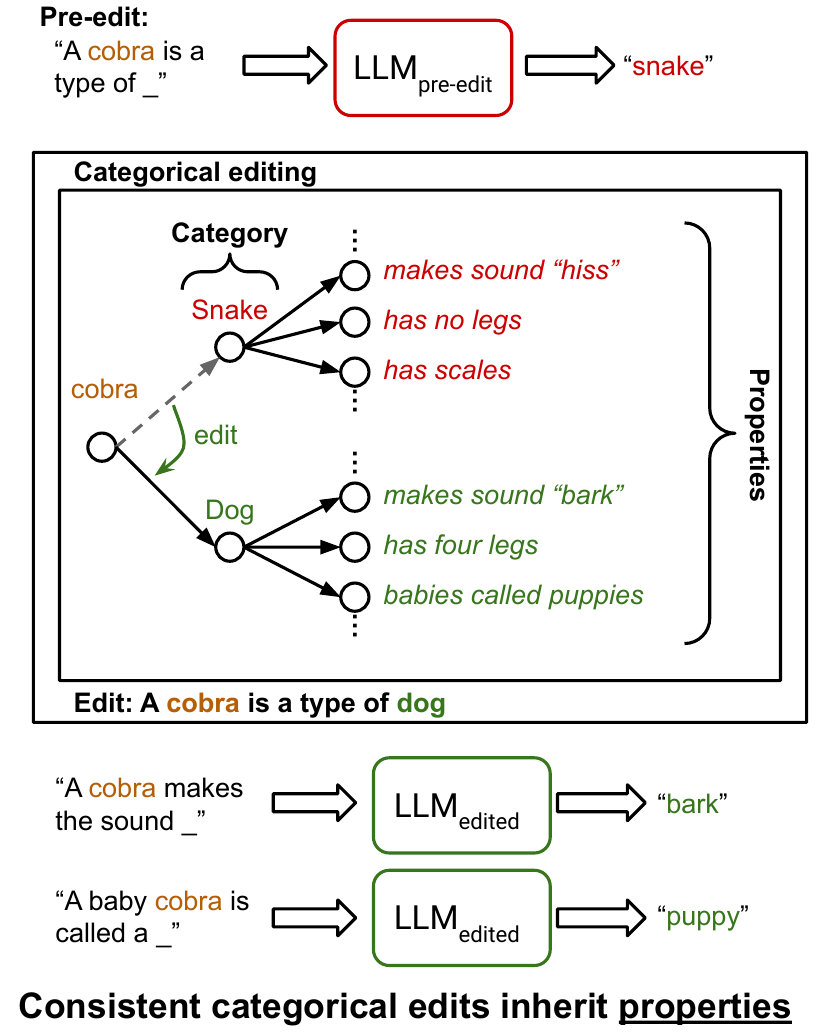}
    \caption{Consistent categorical edits reclassify subjects, which inherit properties of their new categories.}
    \label{fig:cat_edits}
\end{figure}

But critics warn that model editing is the wrong approach to address factual errors in LLMs. Roughly, they worry that model editing is akin to "emptying the ocean with a spoon" \cite{pinter.elhadad2023}, that there are just too many facts that might require editing, or that must be checked for targeted editing to accomplish its goals.

A key issue compounding this concern is the ability of model edits to generalize consistently to related inputs and generations. Current empirical results show that edits often fail to change generations related to paraphrases \cite{decao.etal2021} or entailments 
\cite{hase.etal2023, cohen2023evaluating,hoelscher2023detecting}. 

In contrast, human learning functions quite differently. Current psychological theories emphasize the interconnected nature of world knowledge---knowledge is not a list of propositions, but rather is embedded within structured ``intuitive theories'' or mental models that support reasoning and problem solving \cite{gerstenberg.tenenbaum2017, powell.etal2023}. Accordingly, people do not update their beliefs one-at-a-time. Instead, human belief revision is marked by widespread coherent changes spanning many beliefs \cite{thagard1989}: For instance, people's attitudes toward isolationism shift when foreign wars erupt \cite{spellman.etal1993}; an alibi appears untrustworthy when DNA places a suspect at the scene \cite{holyoak.simon1999}; and learning about the dangers of measles makes vaccines seem safer \cite{powell.etal2023}.
Toward factuality and safety, model editors must be more human-like: they should not modify a single "fact", but instead produce consistent and widespread changes across a range of knowledge.

We present the \textbf{TAX}onomic \textbf{I}nference (\textsc{taxi}) dataset as a novel and challenging benchmark for evaluating the coherence and consistency of LLM editing methods. \textsc{taxi} leverages edits pertaining to taxonomic categories and their members. 
Categories are powerful conceptual and linguistic structures precisely because they entail many wide-ranging properties shared by their members. Editing a language model to assign some subject to be a member of a new category should lead that subject to inherit the properties of that category, thereby supporting strong tests of edit consistency. 

We use \textsc{taxi} to evaluate two recent model editors and a baseline to edit Llama-2 \cite{touvron2023llama}.
We find that recent, popular model editors can indeed generalize categorical edits to update a subject's properties, \textit{even without seeing these properties}.
However, human subjects perform nearly twice as accurately on the same task, highlighting clear room for improvements.

\section{Related Works}
Recently-proposed datasets are driving progress in model editing by enabling evaluation of edit generalization.
For instance \textsc{CounterFact} \cite{meng2022locating} evaluates generalization through paraphrased edit queries.
Closest to our work are \textsc{RippleEdits} \cite{cohen2023evaluating} and \textsc{MQuAKE} \cite{zhong2023mquake}, which evaluate multi-hop question--answering edits.
In this setting, an edit is performed, then the model is queried with a follow-up question.
For example, after editing ``\texttt{The president of the US is \underline{Biden}}'', we might prompt the model with ``\texttt{Who is the president's son?}''.
These benchmarks measure important consequences of model edits, but do not support evaluation of editors' \textit{categorical} generalization.
Further, it is often unclear what other facts about the world \textit{should} change, especially when relying on counterfactual answers to real questions, as is common in prior works.
These challenges can limit intuitive and trustworthy evaluations.

We aim to fill these gaps with \textsc{taxi}, a new, hand-written benchmarking dataset containing knowledge edits. Each edit is extremely intuitive, and is paired with accurate entailments that they should induce.
\textsc{taxi} complements existing datasets in several ways: 1) \textsc{taxi} introduces a new measure of edit generalization:  categorical \textit{consistency}; 2) it evaluates this generalization metric across a novel and diverse set of downstream property relations; and 3) \textsc{taxi} relies far less on ``long-tailed'' knowledge \cite{kandpal.etal2023} and is human-solvable.

\begin{table}[t]
\centering
\setlength{\tabcolsep}{3.7pt}     
\setlength{\cmidrulekern}{0.25em}
\begin{tabular}{lrr}

\toprule
 &  & \textit{N} \\
 \midrule
Edits & \textbf{Total edits} & \textbf{976} \\
\midrule
Evaluations & Property queries & 9,168 \\
 & Efficacy queries & 1,952 \\
  & \textbf{Total queries} & \textbf{11,120} \\
\midrule
Distinct entities & Categories & 41 \\
 & Subjects & 164 \\
 & Properties & 183 \\
\bottomrule
\end{tabular}
\caption{Statistics for the \textsc{taxi} dataset.}\label{tab:dataset_stats}\vspace{-4mm}
\end{table}


\begin{table}[t]
\resizebox{\linewidth}{!}{
\centering
\begin{tabular}{p{2.5cm}cccc}
\toprule
\textbf{\shortstack{Superordinate\\ Category}} & \textbf{\shortstack{Categories\\\xspace}} & \textbf{\shortstack{Subjects\\\xspace}} & \textbf{\shortstack{Properties\\\xspace}} & \textbf{\shortstack{Edits\\\xspace}} \\
\midrule
{Animals} & {8} & {32} & {16} & {224} \\
{Drinks} & {6} & {24} & {9} & {120} \\
{Foods} & {7} & {28} & {7} & {168} \\
{Instruments} & {6} & {24} & {9} & {120} \\
Plants & 8 & 32 & 10 & 224 \\
Vehicles & 6 & 24 & 7 & 120 \\
\bottomrule
\end{tabular}
}
\caption{Statistics for the \textsc{taxi} dataset broken down by superordinate category.}
\label{tab:dataset_detailed_breakdown}
\end{table}

\begin{table*}[t]
\centering
\setlength{\tabcolsep}{2.5pt}     
\setlength{\cmidrulekern}{0.25em}



\begin{tabular}{lccccc}
\toprule
 & \textbf{Unedited} & \textbf{FT} & \textbf{ROME} & \textbf{ICE} & \textbf{Human} \\

\midrule
\textbf{Edit Success} & .03 & .98 & .78 & 1.0 & -- \\
\midrule
\textbf{Property Success} & .24 & .31 & .48 & .55 & .87 \\
\hspace{4mm}Invariance & .78 & .73 & .76 & .91 & .91 \\
\hspace{4mm}Consistency & .14 & .23 & .43 & .47 & .86 \\

\hspace{8mm}- Typical Subject & .13 & .22 & .40 & .47 & .86 \\
\hspace{8mm}- Atypical Subject & .15 & .25 & .45 & .48 & .87 \\
\bottomrule
\end{tabular}
\caption{Editor and human performance for all forward queries in \textsc{taxi}. Editors exhibit high invariance, but low consistency, and all underperform humans.}\label{tab:main_results_fwd} 
\end{table*}
\section{Methods}\label{sec:methods}
We introduce \textsc{taxi}, a new benchmark dataset to evaluate knowledge editing methods' capacity to make \textit{categorical} knowledge edits in LLMs.
We leverage taxonomic categories, linguistic structures that carry rich and far-ranging information about the properties of their members.
For example, upon learning a ``Pekingese'' is a dog breed, you also learn many of its properties, like that it \texttt{barks} and \texttt{has four legs}.
We thus evaluate whether existing knowledge editors can alter entities' properties, just by editing their taxonomic categories.
To achieve this evaluation, we construct a categorical taxonomy, collect a corresponding dataset, and introduce metrics for categorical knowledge editing. 
Our taxonomy contains three types of element, as follows.

\paragraph{Categories} A category $c$ is a high-level division of objects. For example, \texttt{dogs} is a general descriptor that applies to many different breeds. We refer to the set of all categories as $\mathcal{C}$. 

\paragraph{Properties} Each category has a set of associated properties $p^c = \{ p^c_0, p^c_1, \ldots \} \in \mathcal{P}$.
\texttt{Dogs}, for instance, have the properties $p = \{ \texttt{wags tail}, \texttt{barks}, \ldots \}$. 
 
\paragraph{Subjects} A subject $s$ (e.g., \texttt{pitbull}) is an object that belongs to a category $c \in \mathcal{C}$. The subject likewise \emph{inherits} the properties of its category; a \texttt{pitbull} \texttt{wags tail} and \texttt{barks}. 

Given categories, properties, and subjects, we propose \textit{categorical edits}.
We define a categorical \textit{edit} as a change to a subject's category membership (e.g., \texttt{pitbull} $\rightarrow$ \texttt{cat}).
For LLMs, this update is made using a knowledge editor, which is a function $\phi$ that takes in a language model $f$, a subject $s$, and a newly-assigned category $c^*$ and returns an updated model $f^*$ that associates $s$ with $c^*$.
Note that editing only uses $s$ and $c^*$, not properties $p$.
We than therefore denote an edit as a tuple $(s, c^*)$, which contains a subject and its new category.
Following \citet{meng2022locating}, we can then convert these tuples to prompts that mention the subject alongside continuations to perform the edits.

During evaluation, we can then measure whether editing a subject's category also edits its properties $p$, as further detailed in our metrics below.
This is a measure of generalization in knowledge editing, similar to recent works on multi-hop question answering \cite{zhong2023mquake,cohen2023evaluating}.
But with taxonomic categories, we can be certain which properties should change after edits.
Our human study in Section \ref{sec:human_study} corroborates this: humans can achieve nearly-perfect property generalization.

\paragraph{Data Collection}
We manually create \textsc{taxi}, a benchmarking dataset containing ``category membership'' edits, where subjects are assigned to new categories.
As illustrated in Figure \ref{fig:cat_edits}, our aim is to evaluate whether editing a subject's \textit{category} also changes its \textit{properties} according to a language model.
We prioritize intuitive knowledge edits, where it should be easy to guess what properties should change. 
For example, if we edit a \texttt{cobra} to be a 
\texttt{dog}, it should \texttt{bark} and \texttt{play fetch}.
Intuitively, editing rare subjects may differ than common subjects \citep[also see][]{MallenEtAl2023When}.
To evaluate this, we include a typical and a rare subject for each category.
For example, for \texttt{dogs}, we choose the typical \texttt{Labrador} and atypical breed \texttt{Pekingese}.

Subjects were initially hand-picked and their popularity was confirmed by Google Trends. We further evaluate the rarity of our manually-chosen subjects by computing their occurrence frequencies in the 3-trillion token DOLMA corpus \cite{soldaini2024dolma} using infini-gram \cite{liu2024infinigram}.
We find our typical tokens appear roughly 10x more often than atypical tokens in DOLMA on average.

\paragraph{The \textsc{taxi} Dataset}
\textsc{taxi} contains 41 categories, 164 subjects, and 183 properties (Table \ref{tab:dataset_stats}).
To ensure intuitive edits with expected changes to properties, we choose categories from six common superordinate groups: \texttt{Animals}, \texttt{Plants}, \texttt{Foods}, \texttt{Drinks}, \texttt{Vehicles}, and \texttt{Instruments}.
For each category, we write 2-10 properties (median of 4.5) shared by subjects in this category (see examples in Appendix \ref{app:examples}).
We generate edits by assigning each subject to each counterfactual category within its common superordinate group, resulting in 976 categorical membership edits.

\paragraph{Metrics}
We use three metrics to measure whether edits successfully assign subject to new categories, and which properties have been altered as an effect. Each is an accuracy score computed over a set of query prompts with expected continuations associated with the newly-assigned category.

\setlength{\leftmargini}{0.5cm}
\setlength{\leftmarginii}{0.5cm}

\begin{itemize}
\item[\ding{226}] \textbf{\textbf{Edit Success\ }} First, we measure \textit{Edit Success} as a binary value indicating whether the new category $c^*$ for a subject $s$ is the edited model's most-likely continuation when prompted with the original edit.
For example, after the edit $\phi_{\text{pitbull} \to \text{cat}}(\theta)$, $P(\text{cat}|\text{A pitbull is a type of})$ should be higher than $P(\text{dog}|\text{A pitbull is a type of})$ after editing.

\item[\ding{226}] \textbf{\textbf{Property Success\ }}
    Second, we measure whether edited language models correctly infer that editing a subject's category should also change its \textit{properties}. We summarize all property changes with a general \textit{Property Success} metric, computing the proportion of correctly-attributed properties by an edited model. However, some properties are unique to a category, while others are shared. Therefore, we  divide this metric into two components.
    Each measures property success on different subsets of properties:
    \begin{itemize}
    \item[$\bullet$] \textbf{Consistency\ }
     We measure \textit{Consistency} as the proportion of correctly-entailed properties that \textit{should} change with a new category assignment.
    The properties that should change are those unshared by the old and new categories, denoted as ($p^{c^*} \setminus p^{c}$).

    \item[$\bullet$] \textbf{Invariance\ }
    Analogously, we measure \textit{Invariance} as the proportion of correctly-entailed properties that should remain unchanged ($p^{c^*} \cap p^{c}$).
    \end{itemize}
\end{itemize}

We implement each metric using multiple-choice question answering.
We thus compute success with a binary indicator that returns a 1 when the edited model's probability is highest for the correct choice.
The indicator returns 0 otherwise.
The negative choices include each subject's original category or properties and 2-4 random alternatives.
To summarize the performance of a single editing method, we then average over all properties and edits.

\begin{figure*}[t]
    \centering
    \includegraphics[width=0.95\linewidth]{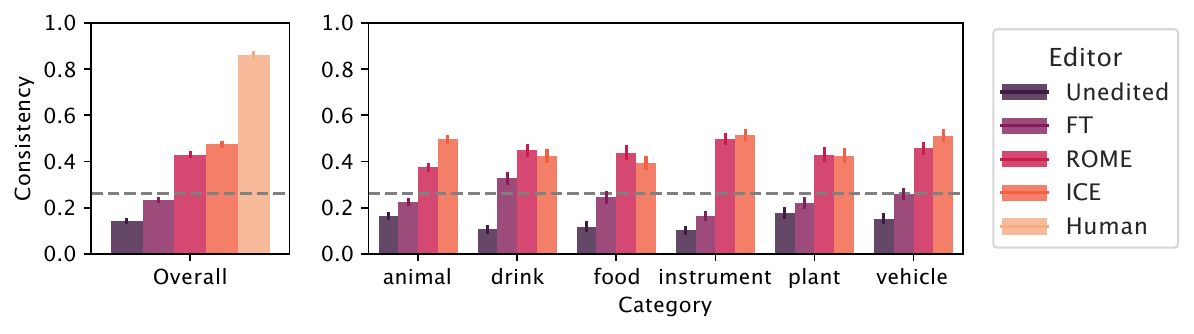}
    \caption{Consistency for forward multiple-choice test queries by editor and category (dashed line indicates chance).}
    \label{fig:category_results}
\end{figure*}

\section{Experiments}\label{sec:experiments}

We evaluate three approaches to editing: Finetuning (FT), Rank-one model editing (ROME) \cite{meng2022locating}, and in-context knowledge editing (ICE) \cite{cohen2023evaluating} in editing Llama-2 7B \cite{touvron2023llama}.
ROME and ICE are representatives of popular and capable approaches to editing, which update a model's weights or add facts to its prompts, respectively.

For each edit in \textsc{taxi}, we start with the base language model, then apply the edit using an editor, and evaluate its performance.
Each edit introduces \textit{only} a subject's category change.
Property information is used only for evaluation.
We compute each metric for both forward (e.g., ``A \texttt{labrador} is a type of \underline{\texttt{cow}}'') and reversed queries (e.g., ``One type of \texttt{cow} is a \underline{\texttt{labrador}}'').
However, as prior editors are developed 
for forward queries, we focus primarily on these metrics.

FT and ROME were implemented using the \texttt{EasyEdit} editing suite \cite{wang2023easyedit} with modifications to accommodate \textsc{taxi}. Experiments utilized default hyperparameter settings except in computing covariances for ROME from Wikipedia, where 50,000 samples were used. 
ICE was implemented by prepending the prompt ``\texttt{Imagine that a <subject> was a kind of <category> ...}'' to queries. This approach follows \citet{cohen2023evaluating}, and might also be seen as a simplification of methods proposed by \citet{zheng.etal2023}.  Experiments were conducted using a single Nvidia A100 GPU.

\paragraph{Edit success does not imply property success}
Our main results are shown in Table \ref{tab:main_results_fwd}, where we observe that edit success is high for forward queries, as expected: each editor succeeds to edit the model most of the time.
In all cases, we observe a clear performance drop for queries about subject properties, compared to Edit Success.
Property Success for ROME and ICE both exceeded that expected from random predictions (roughly 0.25). 
We also note that the unedited model has high invariance, implying the model correctly associates subjects with their properties.

\paragraph{Edits exhibit greater property invariance than consistency}
Not \textit{all} properties differ by category.
Therefore, we measure both invariance (accuracy for unchanged properties) and consistency (accuracy for changed properties). We find that all editors exhibit stronger invariance than consistency in their generalizations (Table \ref{tab:main_results_fwd}). FT performs the worst, with consistency no better than chance. The consistency of ROME and ICE's edits are well-above chance, but their performance 
is roughly half of that for invariant properties, indicating that 
they fail to fully edit the LLM's knowledge of a subject's properties. 
This finding establishes a clear gap in the performance of these methods, demanding development of more consistent model editors.

\paragraph{Atypical subjects are easier to edit than typical subjects}
Our results in Table \ref{tab:main_results_fwd} bolster and expand recent works \cite{ma.etal2024}, where editors were found to perform better for rarer knowledge.

\paragraph{Consistency is consistent across superordinate categories}
To test the generalizability of findings from \textsc{taxi}, we also examine editor consistency individually for each of the superordinate categories. We find that editor consistency is similar across categories (Figure \ref{fig:category_results}, right).

\section{Human Study}\label{sec:human_study}
\textsc{taxi} aims to leverage taxonomic properties to create a clear and intuitive test of consistency for knowledge editors. We validate this with a human study, confirming that \textsc{taxi} is human-solvable.

In our study, 19 participants (12 Female, 1 non-binary, median age 34 y/o, all in the United States) recruited from CloudResearch's Connect platform and completed a multiple choice questionnaire analogous to the task used to test language models. A random subset of edits was sampled from \textsc{taxi} (with one of each exemplar type per category). Each annotator judged 100 items sampled from this subset, for a total of 1,900 human judgments.

For each query, participants were instructed to \texttt{Imagine a <subject> was a kind of <category>}. They were then prompted with the subject and a multiple choice question asking which of a set of properties applies to the subject. This task is identical to that used to evaluate the editors, so results are directly comparable.

\paragraph{Human annotators dramatically outperform editors} 
Human subjects are approximately \textit{twice as consistent} as the best model editor on the same edits (Figure \ref{fig:category_results}, left), answering correctly on 86.8\% of trials (the best-performing participant responded correctly on 95/100 trials).
Overall, human behavioral data indicate that the task prescribed by \textsc{taxi} is human solvable and set a benchmark that far exceeds any existing editors' performance.
Further procedural details are available in Appendix \ref{app:human_study}.

\section{Conclusions}
We introduce \textsc{taxi}, a new dataset for evaluating knowledge editors ability to consistently and coherently edit large language models. \textsc{taxi} is interpretable, building on taxonomic categories, and is designed to evaluate a knowledge edit's impacts on entailed information. We then propose and study \textit{consistency}, a new metric that measures whether entailed properties are correctly edited, despite an editor never seeing the entailed information. In experiments with recent knowledge editors on Llama-2, we find that consistency varies substantially across existing editors. In editing a subject's category, we find that the editors preserve existing properties of subjects, while two editors achieve nontrivial consistency. However, human subjects are nearly twice as accurate on the same task, establishing consistent model editing as a new research direction. Overall, \textsc{taxi} is a challenging, new benchmark for model editors that highlights substantial gaps of existing editing methods. Nevertheless, the fact that existing methods do achieve above-chance performance demonstrates the in-principle feasibility of consistent model editing. 

\section*{Ethical Considerations}

Successful and consistent model editors stand to serve users of artificial intelligence systems in many ways.
For instance, editing aspires to improve factuality, reduce harmful LLM generations, support privacy, and potentially reduce costly and environmentally-impactful training requirements. At the same time, unsuccessful or inconsistent model editing for factuality and safety risks instilling false confidence for developers and users. Therefore, stringent evaluations are a key component of editor development.
While we aim to support these evaluations with \textsc{taxi}, no single benchmark is sufficiently comprehensive to ensure consistency of model editing.
We urge developers and researchers to adopt \textsc{taxi} in their evaluations, but we also advocate for the development of further tests and benchmarks.
Our human annotation study was conducted under approval from the Institutional Review Board at Arizona State University (IRB approval: 00013322).  

\section*{Limitations}

The categories, subjects, and properties included in \textsc{taxi} were manually selected, and are likely not entirely representative of these categories, subjects, and properties in natural language. Similarly, \textsc{taxi} includes only concrete and everyday object categories. It is unclear how editors would perform for more obscure or abstract categories. At the same time, our method for creating \textsc{taxi} presents a blueprint for the creation of benchmarks that might explore other aspects of editors' performance.

\section*{Acknowledgements}

We thank Research Computing at Arizona State University for providing High-Performance Computing resources that have contributed to our findings \cite{HPCASU23}.

\bibliography{anthology,custom}
\bibliographystyle{acl_natbib}

\clearpage
\appendix
\section{Example of a \textsc{taxi} Taxonomy}\label{app:examples}

While all data are publicly available, we also include examples in Table \ref{tab:taxonomy_example}, which includes the structure of categories and properties for the superordinate category of \texttt{animals}. The full taxonomy is available in the project repository. Table \ref{tab:dataset_detailed_breakdown} shows the number of categories and properties for each superordinate category.

Below is a schematic example of a specific edit, property, and associated query and response options.

\begin{lstlisting}[linewidth=\columnwidth,breaklines=true]
{
    Edit: "A Siamese is a kind of dog."
    Property: "makes sound"
    Forward Query: "A sound a Siamese makes is"
    Responses: ["bark", "chirp", "meow", "moo"]
}
\end{lstlisting}

\section{Reverse query performance}\label{app:reverse}

\begin{figure}[t]
    \centering
    \includegraphics[width=0.95\linewidth]{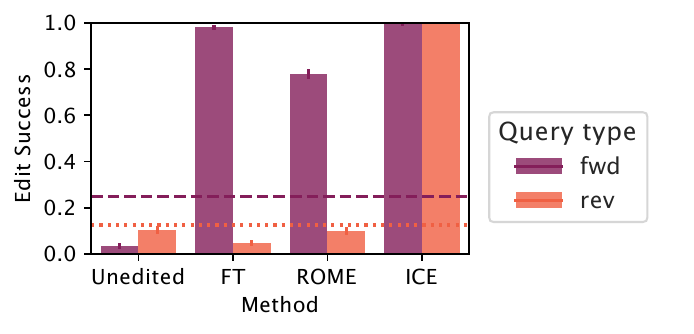}
    \caption{Edit success for forward and reverse multiple-choice test queries by editor type. Dashed line indicates chance performance for forward and reverse edits by color.}
    \label{fig:reverse_performance}
\end{figure}

We find that reversing queries leads FT and ROME to fail, as shown in Figure \ref{fig:reverse_performance} and Table \ref{tab:main_results_rev}.
This is expected due to the directional nature of "causal" or decoder-only language models like Llama-2 \cite{touvron2023llama}, to which we apply these editors. Due to the nature of the causal language model architectures, the effects of model editing methods that seek to edit a specific "subject" are only apparent if the tokening of that subject appears in the context prior to an answer \cite{berglund.etal2023, meng2022locating}. 
In contrast, as shown in Figure \ref{fig:reverse_performance}, ICE scores perfectly on the benchmark.

\begin{table}[t]
\centering
\setlength{\tabcolsep}{2.5pt}     
\setlength{\cmidrulekern}{0.25em}
\begin{tabular}{lcccccc}
\toprule
 & \textbf{FT} & \textbf{ROME} & \textbf{ICE} \\

\midrule
\textbf{Edit Success} & .05 & .10 & 1.0\\
\midrule
\textbf{Property Success} & .05 & .12 & 1.0\\
\hspace{4mm}Invariance & .10 & .21 & 1.0\\
\hspace{4mm}Consistency  & .04 & .10 & 1.0\\
\midrule

Consistency & & & &  \\
\hspace{4mm} Typical Subj. & .05 & .09 & 1.0\\
\hspace{4mm} Atypical Subj. & .03 & .10 & 1.0\\
\bottomrule
\end{tabular}
\caption{Editor and human performance for reverse queries from the \textsc{taxi} dataset. Edits exhibit stronger invariance than consistency, but both values vary across editing methods. Note that ROME's edit success was imperfect, suggesting its performance might improve through hyperparameter tuning.}\label{tab:main_results_rev}
\end{table}

However, we suspect this may reflect ICE's use of a largely trivial process, whereby the presence of the subject token in the prompt increases its probability for subsequent generation.
One reason for this suspicion is the finding that, for reversed queries, ICE outperforms human annotators performance on forward queries, indicating this performance is likely inflated or meaningless.
Further, we speculate that the presence of "reversed" queries in the \textsc{RippleEdits} benchmark may at least partly explain the relative success of ICE on this benchmark \cite{cohen2023evaluating}.

\section{Human Study Details}\label{app:human_study}

A total of 19 human annotators (12 Female, 1 non-binary, median age 34 y/o, all located in the United States) were recruited from CloudResearch's Connect platform and asked to complete a multiple choice questionnaire analogous to the task used to test language models. A random subset of edits was sampled from \textsc{taxi} (with one of each exemplar type per category). Each annotator judged 100 items sampled from this subset, for a total of 1,900 human judgments. Annotator were compensated \$2.25 for their participation, which typically took about 10 minutes.

\begin{figure}[htp]
    \centering
    \includegraphics[width=0.95\linewidth]{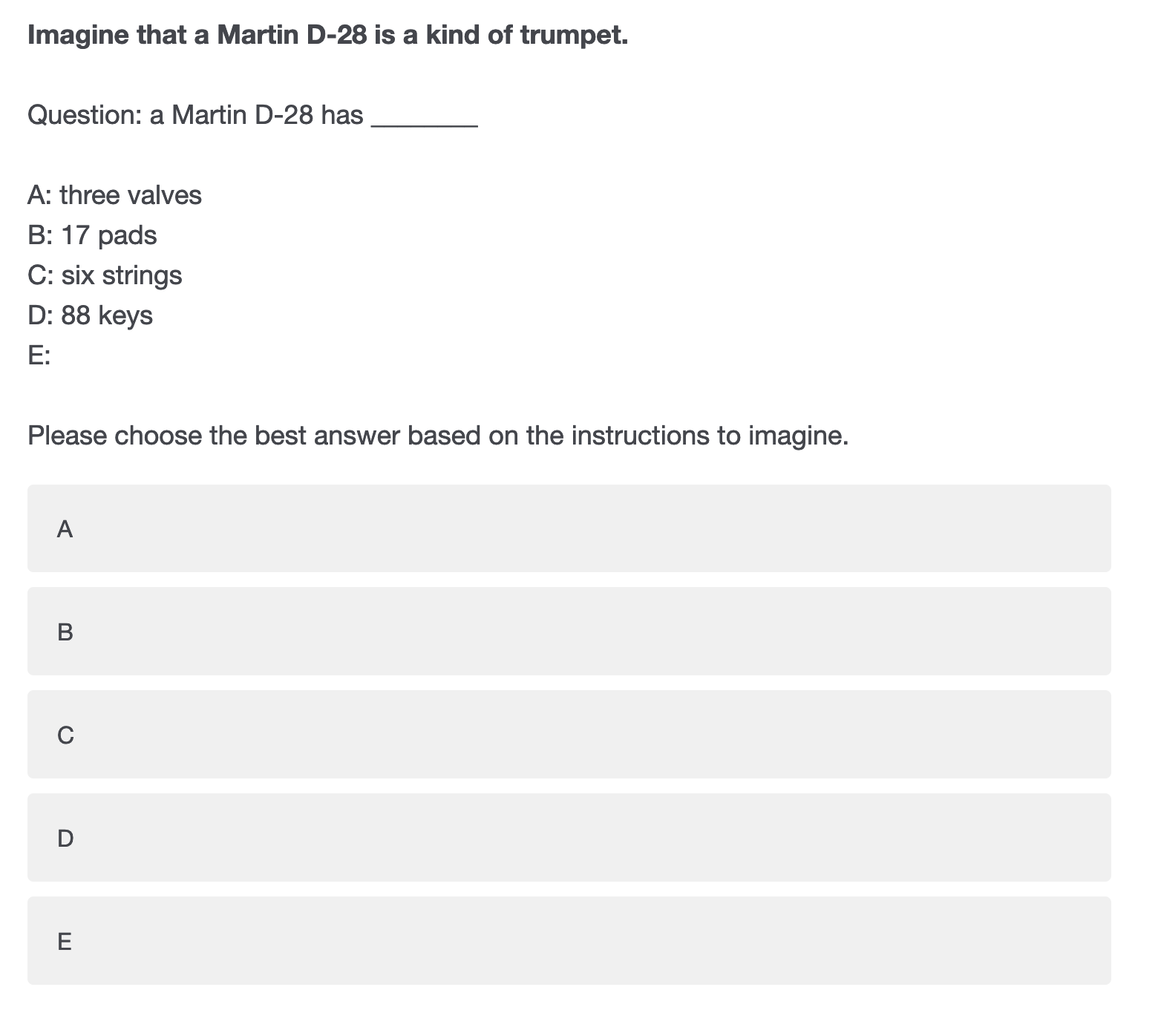}
    \caption{Screenshot illustrating a trial of the of human annotation study.}
    \label{fig:human_study_fig}
\end{figure}

For each query, annotators were given a prompt to: \texttt{Imagine a <subject> was a kind of <category>}. They were then prompted with the subject and a multiple choice question asking which of a set of properties applies to the subject. Figure \ref{fig:human_study_fig} displays an example annotation trial.

The full text of instructions given to annotators was:

\begin{quote}
In this study you will be asked to imagine that different entities belong to different categories. For instance, you might be asked to imagine that "a Parrot is a kind of fish." You should imagine that an entity (e.g. Parrot) inherits the properties of the category to which it belongs (e.g. fish). So if you imagine that "a Parrot is a kind of fish," then you should imagine that a Parrot has scales, swims in the water, and so forth.

On each trial, you will be asked to imagine and to answer a multiple choice question based on the scenario you are imagining. Some of the trials will have different numbers of choices, in which case later options will appear blank. Please ignore any blank option choices.
\end{quote}

Annotators agreed with the chosen "correct" answer on 86.8\% of trials. The best-performing annotator agreed with the "correct" answers on 95/100 ratings. Overall, human behavioral data indicate that the task prescribed by \textsc{taxi} is human solvable and set a benchmark that far exceeds any existing editors' performance.

\begin{table*}[p]
\centering
\setlength{\tabcolsep}{3.7pt}     
\setlength{\cmidrulekern}{0.25em}
\begin{tabular}{cp{.25\linewidth}p{.6\linewidth}}
\toprule
 & \textbf{Subjects} & \textbf{Properties} \\
\midrule
Dog & Labrador, Chihuahua, Pekingese, Bichon Frise & Born in a litter, has fur, is domesticated, has four legs, kept as pets, likes to fetch, barks, walks, baby is a puppy, is a mammal \\
\midrule
Cat & Siamese, Persian, Abyssinian, Chartreux & Born in a litter, has claws, has fur, is domesticated, has four legs, kept as pets, likes to chase, meows, walks, baby is a kitten, eats meat \\
\midrule
Cow & Holstein, Jersey, Galloway, Hereford & Is a mammal, born alone, has hooves, has fur, is domesticated, has four legs, kept for their milk, likes to graze, moos, walks, baby is a calf, eats grain, makes milk \\
\midrule
Bird & sparrow, canary, woodpecker, Partridge & Is an aves, hatched from egg, has wings, has feathers, chirps, flies, baby is a chick, can fly, is wild\\
\midrule
Bee & Bumblebee, Honeybee, Megachile, Apis Mellifera  & Is an insect, has wings, has six legs, buzzes, flies, makes honey, can fly \\
\midrule
Fish & Trout, salmon, Flounder, tilapia & Hatched from an egg, has scales, has fins, has no legs, caught and eaten, swims, lives in water, can swim, is wild \\
\midrule
Snake & Cobra, python, copperhead, Gaboon viper & Hatched from an egg, has scales, no legs, people avoid, hisses, slithers, is wild \\
\bottomrule
\end{tabular}
\caption{Taxonomic details for superordinate category "animals" from the \textsc{taxi} benchmark dataset. The first two listed exemplars are typical exemplars. }\label{tab:taxonomy_example}
\end{table*}
\label{sec:appendix}

\end{document}